\documentclass[twocolumn, switch]{article} 

\usepackage{preprint}
\usepackage{subcaption}
\usepackage[export]{adjustbox}

\usepackage{amsmath, amsthm, amssymb, amsfonts}
\usepackage{tabularray}

\usepackage[numbers,square]{natbib}
\usepackage[utf8]{inputenc}	
\usepackage[T1]{fontenc}	
\usepackage{xcolor}		
\usepackage[colorlinks = true,
            linkcolor = purple,
            urlcolor  = blue,
            citecolor = cyan,
            anchorcolor = black]{hyperref}	
\usepackage{booktabs} 		
\usepackage{nicefrac}		
\usepackage{microtype}		
\usepackage{lineno}		
\usepackage{float}			

\usepackage{lipsum}		

\usepackage{newfloat}
\DeclareFloatingEnvironment[name={Supplementary Figure}]{suppfigure}
\usepackage{sidecap}
\sidecaptionvpos{figure}{c}

\usepackage{titlesec}
\titlespacing\section{0pt}{12pt plus 3pt minus 3pt}{1pt plus 1pt minus 1pt}
\titlespacing\subsection{0pt}{10pt plus 3pt minus 3pt}{1pt plus 1pt minus 1pt}
\titlespacing\subsubsection{0pt}{8pt plus 3pt minus 3pt}{1pt plus 1pt minus 1pt}

\usepackage{tikz}

\definecolor{lime}{HTML}{A6CE39}
\DeclareRobustCommand{\orcidicon}{
	\begin{tikzpicture}
	\draw[lime, fill=lime] (0,0) 
	circle [radius=0.16] 
	node[white] {{\fontfamily{qag}\selectfont \tiny ID}};
	\draw[white, fill=white] (-0.0625,0.095) 
	circle [radius=0.007];
	\end{tikzpicture}
	\hspace{-2mm}
}
\foreach \x in {A, ..., Z}{\expandafter\xdef\csname orcid\x\endcsname{\noexpand\href{https://orcid.org/\csname orcidauthor\x\endcsname}
			{\noexpand\orcidicon}}
}

\newcommand{\E}{\mathbb{E}}

\newcommand{\N}{\mathcal{N}}
\renewcommand{\u}{\mathbf{u}}

\newcommand{\x}{\mathbf{x}}
\newcommand{\z}{\mathbf{z}}
\renewcommand{\r}{\mathbf{r}}
\newcommand{\s}{\mathbf{s}}
\renewcommand{\v}{\mathbf{v}}

\newcommand{\p}{\mathbf{p}}

\usepackage{algorithm,algorithmic}
\usepackage{enumitem}

\title{A Variational Autoencoder Framework for Robust, Physics-Informed Cyberattack Recognition in Industrial Cyber-Physical Systems}

\usepackage{eso-pic,picture}

\AddToShipoutPicture{%
\setlength{\unitlength}{1pt}%
\protect{\put(.5\paperwidth,.5\paperheight){%
\makebox(0,-745){\rotatebox{0}{\textcolor[gray]{0.60}%
   {\color{gray}{*correspondence: \texttt{dli4@clemson.edu}}}}}
\makebox(0,755){\rotatebox{0}{\footnotesize\textcolor[gray]{0.60}%
   {\color{gray}{This work has been submitted to the IEEE for possible publication.}}}}

\makebox(0,735){\rotatebox{0}{\footnotesize\textcolor[gray]{0.60}%
   {\color{gray}{Copyright may be transferred without notice, after which this version may no longer be accessible.}}}}
   
   }}\ClearShipoutPictureBG}

\usepackage{abstract}
\usepackage[affil-it]{authblk}

\author[1\thanks{\tt{naftabi@clemson.edu}}]{Navid Aftabi}
\author[1\thanks{\tt{dli4@clemson.edu}}]{Dan Li, Ph.D.}
\author[2\thanks{\tt{paritosh.ramanan@okstate.edu}}]{Paritosh Ramanan, Ph.D.}

\affil[1]{Department of Industrial Engineering, Clemson University, Clemson, SC, USA}
\affil[2]{School of Industrial Engineering and Management, Oklahoma State University, Stillwater, OK, USA}

\begin{document}

\twocolumn[ 
\begin{@twocolumnfalse} 
  
\maketitle

\begin{abstract}
Cybersecurity of Industrial Cyber-Physical Systems is drawing significant concerns as data communication increasingly leverages wireless networks. A lot of data-driven methods were develope for detecting cyberattacks, but few are focused on distinguishing them from equipment faults. In this paper, we develop a data-driven framework that can be used to detect, diagnose, and localize a type of cyberattack called covert attacks on networked industrial control systems. The framework has a hybrid design that combines a variational autoencoder (VAE), a recurrent neural network (RNN), and a Deep Neural Network (DNN). This data-driven framework considers the temporal behavior of a generic physical system that extracts features from the time series of the sensor measurements that can be used for detecting covert attacks, distinguishing them from equipment faults, as well as localize the attack/fault. We evaluate the performance of the proposed method through a realistic simulation study on a networked power transmission system as a typical example of ICS. We compare the performance of the proposed method with the traditional model-based method to show its applicability and efficacy.
\end{abstract}
\def\abstractname{Note to Practitioners}
\begin{abstract}
    To combat the cybersecurity threats that networked industrial control systems are facing, we developed a cyberattack detection and localization method that uses machine learning that fuses the domain knowledge with information extracted from the sensor data to identify the occurrence and the location of a covert cyberattack in a robust manner. This is fulfilled by merging the data features representing the change in structural correlation, temporal correlation, and physical behavior, which are extracted based on VAE, RNN, and state estimation, respectively. By using the developed data-driven framework, practitioners enable timely and accurate cyberattack detection and location identification in networked industrial control systems. This provides critical insights in post-detection investigation and attack recovery. Moreover, the proposed framework is robust to attack data in the training data, which reduces the risk of misclassification when the historical data available are not accurately labeled.
\end{abstract}

\vspace{0.35cm}

\end{@twocolumnfalse} 
] 
\saythanks



\section{Introduction}
Industrial control systems (ICS) are foundational components geared towards control and monitoring in power, energy, chemical and manufacturing domains \cite{stouffer2011guide}. The increased embedding of IoT sensors across ICSs in recent times has resulted in significant amounts of monitoring and operations-oriented sensor data being collected in real-time. While the integration of IoT sensors presents a remarkable opportunity for optimizing and streamlining system operations, it also raises cybersecurity concerns with respect to the underlying infrastructure assets \cite{desmit2017approach,sridhar2011cyber}. Being inherently affiliated with the Cyber-Physical Systems (CPS) domain, ICSs are characterized by a cyber layer that continuously interacts with the physical layer components. The cyber layer generates data-driven insights and decisions that are applied to the underlying physical components. Meanwhile, the sensors on the physical components acquire real-time operational data which is relayed back to the cyber layer components. Due to the control-oriented interdependencies between the cyber and the physical layers, malicious data integrity attacks can cause catastrophic damage to the underlying physical assets. Such attacks alter sensor measurements and control actions at the level of the programmable logic controllers (PLCs) resulting in complete ignorance of the true physical behavior of assets at the ICS level. For instance, the onset of the \textit{Stuxnet} \cite{kushner2013real} malware in 2010 that inflicted data integrity attacks, demonstrated the magnitude of physical damage wreaked by such attacks on critical infrastructure systems. Consequently, data integrity attacks pose an important challenge in terms of IoT-enabled ICSs \cite{sridhar2010data}.

\subsection{Related Work}
Handling data integrity attacks primarily involves two main steps related to detection and diagnosis. Detection of data integrity attacks pertains to the task of ascertaining anomalous behavior during the operation of physical assets. Largely, detection methodologies can be classified into either model-based or data-driven categories. Model-based methods require engineering domain knowledge in order to establish physical rules that might govern sensor measurements under normal circumstances \cite{ding2020secure,liu2011false,van2015sequential, cardenas2011attacks,huang2018online,li2014quickest}. Therefore, these methods detect attacks by identifying anomalies with respect to the established baseline stemming from the physical rules. A prevalent model-based rubric is to leverage observed sensor measurements and compute residuals with respect to the estimated sensor measurements \cite{van2015sequential,cardenas2011attacks}. Anomalies can then be detected by exploiting the fact that under routine operations, residuals are expected to be normally distributed around a mean of zero. On the other hand, a large residual implies a greater discrepancy between the observed and expected measurements indicating anomalous behavior. The detection of large residuals can be accomplished by using the $\chi^2$ detector, which raises an alarm when the sum of squared residuals (SSR) is above a certain threshold that is determined through engineering knowledge. Model-based methods have an inherent drawback due to the requirement of an accurate physical model that accurately represents a complex set of spatial and temporal interdependencies in a system comprising of multiple subsystems. In case of power systems, the system is usually represented by a steady-state power flow equations that rely on independent observations as input without considering the time-varying system dynamics.

In contrast to model-based approaches, data driven methods sift through the sensor measurements and control inputs and outputs in order to learn ML or analytical models that can represent the normal behavior of the system without necessarily understanding the physics behind such behavior. Consequently, data-driven approaches offer much more flexibility as a result of eliminating the need to incorporate physical rules. Instead, these methods rely on inferring the abnormal behaviors associated with equipment performance that is implicit in the sensor data. 

A popular category of data driven detection mechanisms pertains to intrusion detection systems \cite{yilmaz2017mitigating,terai2017cyber,caselli2016specification}. These methods focus on analyzing network traffic data in order to detect intrusions \cite{yilmaz2017mitigating,terai2017cyber,caselli2016specification}. While they provide a good litmus test for detecting cyber intrusions, they are ineffective for detecting anomalous behavior on signals stemming from the underlying physical processes associated with the assets \cite{li2020deep}. The inability to apply intrusion detection methodologies largely stems from the stable and consistent behavior exhibited by assets in the short term that only exhibits degradation over a much longer time horizon. Therefore, there have been attempts to leverage recurrent neural networks (RNNs) \cite{li2020deep} and convolutional neural networks (CNNs) \cite{bakalos2019protecting}, in order to enable the fusion of traffic and sensor data to learn normal system behavior. Such methods have been successfully demonstrated on control systems to characterize physical intrusions on a single ICS. However, most of these methods are not capable of handling subsystem level interaction typically found in networked ICSs wherein a single utility provider owns and operates multiple subsystems. Consequently, it is imperative to develop data driven methodologies in order to detect cyberattacks on networked ICSs. 

There are a few drawbacks in using data-driven methods for cyberattack detection in ICSs. First, data-driven anomaly detection methods may suffer from false alarms triggered by natural equipment faults \cite{buczak2015survey}. False alarms are detrimental to the smooth functioning of the networked system by causing unnecessary shutdowns and inspections that could drastically increase the overall operations cost of the entire network. Another layer of complexity is introduced by the high degree of interconnectivity present in system of subsystems, wherein a data integrity attack on one asset might implicate several other components of the networked infrastructure. The key to reducing false alarms lies in being able to diagnose anomalies yielded by data-driven detection methods. A classical technique to diagnose anomalies leverages a human-in-the-loop system that requires a domain expert to periodically classify anomalies as they appear. Such techniques to diagnose anomalies can create a critical bottleneck owing to reduced scalability and the degree of manual intervention needed. Therefore, recent advances have focused on deep-learning methods in order to achieve scalability and eliminate manual interventions \cite{balta2023digital,zhang2019multilayer,al2020ensemble,li2022online}. Another challenge that hamstrings deep learning methods is the lack interpretable diagnosis results. In other words, most deep-learning based methods restrict themselves to pure classification tasks without delving deeper into quantifying the uncertainties associated with the predictions. Quantifying these uncertainties is vital to cyberattack recognition since it provides significant context to enable decision-making with respect to responding to alarms as they occur. 

\subsection{Challenges and Contributions}
\begin{figure*}[!ht]
\centering
\includegraphics[width=.99\textwidth]{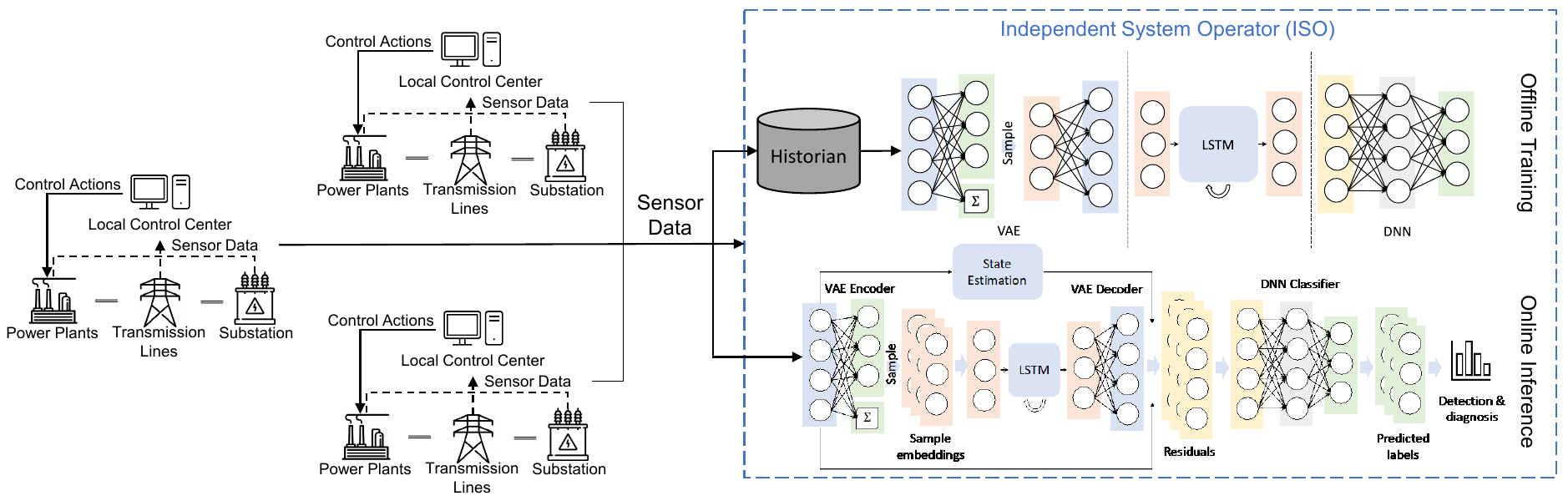}
\caption{The detection and identification architecture}
\label{fig:arch}
\end{figure*}

In order to address the aforementioned drawbacks, in this paper we develop a holistic data driven detection and diagnosis scheme. Our approach incorporates the salient features of model-based and data-driven methods while quantifying the uncertainties associated with diagnosis tasks. As shown in Figure \ref{fig:arch}, we consider a transmission system comprising multiple local control centers (LCCs) regulating regional power generation in coordination with an independent system operator (ISO). The ISO is in charge of capturing sensor data from LCCs for state estimation based on which optimum operational schedules can be determined. In conjunction with the ISO, the LCCs dynamically control the network. 

We consider cyberattacks that manipulate sensor data and control actions simultaneously at the LCC level so as to disrupt power generation. While it is possible to use dynamic state estimation to detect false data injection attacks \cite{kazemi2020secure, chakhchoukh2019diagnosis}, such methods are ineffective when considering replay attacks and covert attacks locally \cite{de2017covert,li2022online,romagnoli2019model}. Designing a detection scheme at the ISO level requires us to characterize the correlation in system behavior among different LCCs in case of an attack, which is particularly challenging for the following reasons: First, the complexities stemming from dependencies in LCC operations poses challenge in building an accurate model at scale. On the one hand, data-driven detection methods tend to suffer from performance degradation over time and as the system scale goes up. On the other hand, model-based detection methods relying on physical state estimation may be ineffective in detecting certain types of attacks due to their lack of flexibility. Therefore, a physics-informed machine learning model needs to be designed to achieve a high performance in detecting and diagnosing attacks for a large scale ICS network. Second, the historical dataset collected in practice are imbalanced and subject to unidentified attack data. It is common that the majority of the observations in the dataset are labeled as normal, and the data labeled normal may also contain attack data. While data imbalances can be properly solved by down sampling and upweighting, the contaminated normal dataset can cause an anomaly detection model to suffer from significant false alarm and misdetection rates.

The crux of our approach relies on combining data-driven and model-based approaches to build a physics-informed machine learning framework in order to characterize the system correlations and detect and recognize attacks in a robust manner. Specifically, we use a Variational Autoencoder, an LSTM, and a DNN as the drivers for the data driven approach whereas, the model-based aspect is handled by the classical state estimation tools. The uniqueness of our approach stems from the fact that the spatial correlations can be captured using the VAE while the temporal behavior is learnt using LSTMs. These steps are followed by a classification task conducted using DNNs that are reinforced using outputs of the state estimation. Our pipelined framework enables ISOs to continuously monitor and learn system behavior by accomplishing \textbf{both} attack detection and diagnosis. Our specific contributions that differentiates our work from some existing VAE-based attack detection works \cite{fahrmann2022lightweight, takiddin2021variational,zhang2022multiclass} are summarized as follows:
\begin{itemize}
    \item We develop a centralized physics informed cyberattack recognition framework capable of operating at the aggregator level. The framework integrates the advantages of model-based and data-driven methods, where it identifies anomalous events with high accuracy by incorporating system physics.
    
    \item We demonstrate that using VAE and LSTM, the spatial and temporal correlations in the sensor data can be captured effectively. As a result, we show that our framework improves robustness of detection even in the event of misclassified training data.
    
    \item We develop a voting mechanism coupled with the VAE and the DNN classifier, which approximates a probability distribution that quantifies the likelihood of anomalous events. This enables uncertainty quantification that facilitates decision making with respect to diagnosis. 
\end{itemize}

The rest of the paper is organized as follows: Section \ref{sysmodel} introduces the generic model of networked industrial control systems; Section \ref{method} covers the proposed framework by first introducing the overall architecture, and then providing technical details of the offline training and online inference algorithms; Section \ref{performance} covers the performance evaluation, where we compare the performance of the proposed method with baseline methods and state-of-the-art techniques. Conclusions and future directions are provided in Section \ref{concl}.

\section{System Model}
\label{sysmodel}
\subsection{Networked ICS}
We consider a networked control system consisting of $k$ subsystems as shown in Fig. \ref{fig:arch}. The subsystems are not necessarily homogeneous, meaning there is not a single model that can represent the behaviors of all the subsystems. For example, a smart grid is a network of generation buses, representing the power plants, and load buses, representing the substations, and different models should be used to represent the power plants and the substations. We denote the state of the subsystem $i$ at time $t$ by a vector $x^i_t$, where $x^i_t \in R^{n_i}$, and $n_i$ is the minimum number of variables needed to uniquely define the system state of site $i$. Denote the history of state of subsystem $i$ till time $t$ as $x^i_{\{t\}}=\{x^i_0,x^i_1,...,x^i_t\}$. The full system state vector at time $t$, $x_t$, is given by a concatenation of all the subsystem states. i.e, $x_t^T=[x^{1T},...,x^{kT}]\in R^{N}$ where $N=n_1+n_2+...+n_k$. Note that $x$ is not directly observed, but inferred from the measurements from the $M$ ($M>N$ for system observability) sensors distributed throughout the system. Similarly, we have $y_t^T=[y^{1T},...,y^{kT}]\in R^{M}$ where $M=m_1+m_2+...+m_k$, and $m_i$ is the number of sensors in subsystem $i$. Denote the control action as $u_t^T=[u^{1T},...,u^{kT}]\in R^{P}$, where $u^i_t\in R^{p_i}$ and $P=p_1+...+p_k$. Similarly, we denote the history of control actions and measurements till time $t$ as $u_{\{t\}}=\{u_0,u_1,...,u_t\}$ and $y_{\{t\}}=\{y_0,y_1,...,y_t\}$, respectively.

In general, the dynamics of the networked system can be represented by
\begin{gather}
    x_t=\mathcal{F}(x_{t-1},u_{t}),\label{ngs1}\\
y_t=\mathcal{G}(x_t),\label{ngs2}
\end{gather}
and a single subsystem $i$ can be represented by a recursive function of $x^i_t$ as follows:
\begin{gather}
    x^i_t=f_i(x_{t-1},u_{t-1})\label{gs1}\\
y^i_t=g_i(x_t)\label{gs2}
\end{gather}
Systems with linear dynamics where the Markovian property holds, can be represented by a linear state-space model. For an individual subsystem $i$ which operates independently, we have:
\begin{gather}
x^i_t=A_ix^i_{t-1}+B_iu^i_{t-1}\label{ss1}\\
y^i_t=C_ix^i_t\label{ss2}
\end{gather}

When the subsystems are correlated with each other, the full system can be modeled by
\begin{gather}
x_t=Ax_{t-1}+Bu_{t-1},\\
y_t=Cx_t,
\end{gather}
where $A$, $B$, and $C$ are functions of $A_i$'s, $B_i$'s, and $C_i$'s. In model-based methods, typically the above model is used to build a state estimator $\mathcal{E}$ (e.g., a Kalman filter or a first order state estimator) that estimates $\hat{x}_t$ based on the observed measurement $y_t$ and the previous estimation $\hat{x}_{t-1}$. The estimated $\hat{x}$ is then substituted back into the system model to predict the next measurement $\hat{y}_{t+1}$, which is compared with the observed $y_{t+1}$ to obtain the residuals $r_t$:
\begin{gather}
\hat{y}_t= C\mathcal{E}(y_{t-1},\hat{x}_{t-1}),\\
r_t=y_t-\hat{y}_t.
\end{gather}

When accurately parameterized, the above physics-based model works well for simple systems. However, one disadvantage of these physics-based models is the estimation of all the parameters requires a large amount of data, and might cause identifiability issues. Therefore, We assume the interconnectivity of these systems are complex to represent using a physics-based model, this is true especially when the system is nonlinear and the subsystems are interconnected. Another disadvantage of physics-based models is that for larger networked ICSs like power systems, the state estimation is only based on the measurement function in \eqref{ngs2} without considering the dynamic behavior of the subsystems or the temporal correlation characterized by \eqref{ngs1}. Note that the state evolution also depends on the control input $\u$. In practice, the control actions are calculated by solving an optimization problem based on a period of historical data, or on a state estimation from the previous time step. In this case, the temporal correlation is implicit and hence challenging to incorporate into the model or the state estimator. 
Nevertheless, detecting cyberattacks often hinges on temporal correlation. This is because the commencement of such attacks can cause unusual short-term system behaviors. These behaviors might elude the spatial correlation defined by the measurement function, but they become more apparent when analyzed through a temporal lens.

As an example, in a power system, the state vector $x$ contains the control actions $u_t$ are the power generation setpoints of all the generators. The setpoints are calculated using the well known Mixed Integer Unit Commitment (MIUC) \cite{carrion2006computationally} formulation given in Problem \eqref{eq:centUC}.
\begin{subequations}\label{eq:centUC}
\begin{align}
\qquad & \underset{\alpha,\theta}{\text{min}} & c^\top \alpha+ d^\top \theta \label{eq:centUC0}&&\\
& \text{subject to} & Q\theta + R \alpha= E  &  \label{eq:centUC1}&\\ 
& &F \theta = H & \; \; \; & \label{eq:centUC2}
\end{align}
\end{subequations}
are the power generation setpoints of all the generators. As mentioned earlier, the implicit temporal correlation generated by calculating the control actions in a history-dependent manner as in \eqref{eq:centUC} cannot be easily represented by any physics-based model. On the other hand, the occurrence of an attack might be easily captured by analysing the temporal behavior of the system, while the steady state does not show any anomaly. Therefore, data driven method, specifically RNN, well fits this situation where temporal correlation needs to be captured via data-driven techniques.

\begin{figure*}[!ht]
    \centering
    \includegraphics[width=0.72\textwidth]{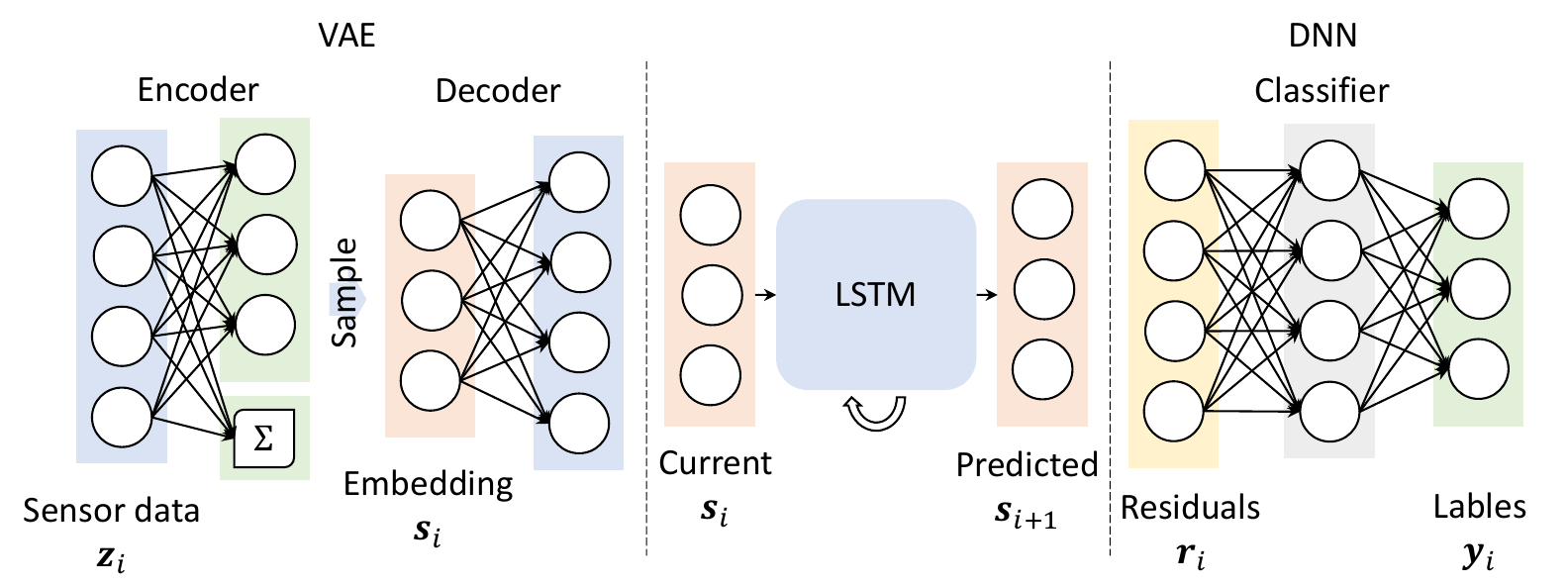}
    \caption{The offline training framework}
    \label{fig:offline}
\end{figure*}

\subsection{Covert Attack Model}
The covert cyberattack was first proposed in \cite{smith2011decoupled} for a single linear system as represented by functions \ref{ss1} and \ref{ss2}. Specifically, the attacker knows the values of $B_i$ and $C_i$. Recall that a covert attacker is assumed to possess access to control action and the sensor measurements as well. Under these assumptions, the attacker implements the covert attack by the following steps:
\begin{itemize}
    \item  First, the attacker manipulates the control actions using the following equation
    \begin{equation}
        \tilde{u}^i_t=u^i_t+a_t \label{ca1},
    \end{equation}
    where $\tilde{u}^i_t$ is the manipulated control action at time $t$, $a_t$ is the attack signal added to the original control action $u^i_t$. According to \ref{ss1}, this manipulation will alter the system state at $t+1$ by $B_ia_t$. That is,
    \begin{equation}
        \tilde{x}^i_{t+1}=x^i_t+B_ia_t 
    \end{equation}
    The consequent sensor measurements will be biased by $C_iB_ia_t$. That is,
    \begin{equation}
        \tilde{y}^i_{t+1}=y^i_{t+1}+C_iB_ia_t .
    \end{equation}
    \item Then, the attacker manipulates the sensor measurements by subtracting the above bias. i.e.,
    \begin{equation}
        \check{y}^i_{t+1}=\tilde{y}^i_{t+1}-\gamma,
    \end{equation}
    where $\gamma=C_iB_ia_t$. In this way, the manipulated measurement $\check{y}^i_{t+1}$ is equal to the expected measurement $y^i_{t+1}$ without attack. Hence, the covert attack can be successfully disguised, as the measurement is the only output of the system, which means all the data inference is conducted based on $y$.
\end{itemize}

In this work, we generalize the covert attack to nonlinear systems represented by Equations (\ref{gs1}) and (\ref{gs2}). We assume the attacker gains knowledge of the dynamics of subsystem $i$. This could be taken as the attacker obtains an estimation of the local functions $\hat{f}_i$ and $\hat{g}_i$ in Equations (\ref{gs1}-\ref{gs2}), which can serve as the simulator of system $i$. With this knowledge as well as the access to the control actions and sensor measurements, the attacker conducts a covert attack using the following steps:
\begin{itemize}
    \item  First, the attacker reads the original control action $u_t$ and simulates the expected sensor measurements using the knowledge of subsystem $i$. That is,
    \begin{gather}
        \hat{x}^i_{t+1}= \hat{f}_i(x_{t},u^i_{t}),\\
        \hat{y}^i_{t+1} = \hat{g}_i(\hat{x}^i_{t+1}).
    \end{gather}

    \item Then, the attacker manipulates the control actions as in Eq. (\ref{ca1}). According to Eq. (\ref{gs1}), this manipulation will alter the system state at $t+1$ as well as the sensor measurements. That is,
    \begin{gather}
        \tilde{x}^i_{t+1}=f_i(x_{t},\tilde{u}^i_{t},)\\
        \tilde{y}^i_{t+1} = g_i(\tilde{x}^i_{t+1}).
    \end{gather}

    \item Finally, the attacker replaces the consequent sensor measurements with the simulated one. i.e.,
    \begin{equation}
        \check{y}^i_{t+1} \leftarrow \hat{y}^i_{t+1}.
    \end{equation}
\end{itemize}
Notice that we assume the attacker's access to the sensors is limited to subsystem $i$. This means the attacker is not capable of compensating for the impact of attacking subsystem $i$ on other subsystems. This fact lays the foundation for our detection and localization framework. However, detecting the attack in this case is nontrivial because the sensors that are most informative of the attack are manipulated, while the attack's impact on other sensors does not have a clear indication of the occurrence of an attack as well as its location.

\section{Methodology}\label{method}
\subsection{The Overall Architecture}
The proposed deep learning architecture is shown in Fig.\ref{fig:arch} in the context of power transmission systems. In a networked ICS, the local control center at each region collects the local sensor data and makes power generation control decisions for the generators locally. The aggregator (independent system operator) estimates the global system state, makes power generation plans, and distribute the plans in terms of the power generation setpoints to the local control centers. The aggregator has access to all the sensor data, while each local control center only have access to the local sensor data. Our algorithm is implemented at the aggregator level, which contains two modules: the offline training and online inference module. The offline training module involves training three components: a variational autoencoder, an LSTM network, and a deep neural network. In the training phase, the three components are trained sequentially. The trained networks will then be reorganized and used in the online inference module.

\subsection{Offline Training}
In the offline training stage, a VAE, RNN, and a DNN are trained separately. The VAE and RNN are trained using normal historical data that potentially contain mislabeled attack data, and the DNN is trained using a mixture of labeled attack and normal data. Below we elaborate the training details for each of the compoenent.

\textbf{VAE:} Autoencoders are widely used to extract the lower dimensional embedding of high dimesional data in an unsupervised manner. Compared to classic autoencoders, VAE is more robust to bad data as because of the sampling step between the encoder and the decoder, getting more research attention in anomaly detection and diagnosis \cite{oliveira2023early,ibrahim2023non}. The variational autoencoder is trained on the sensor data under normal conditions. The sensor data include the power flow collected from the transmission line $\p_f$, the production on each generator $\p_g$, and the voltage measurements on each bus $\v$. The concatenated sensor measurement samples $\z_i=(\p_f,\p_g,\v)_i$, $i=1,...,N$ are used to train the VAE. The VAE consists of an encoder that maps the input sensor data to the latent space of a probability distribution of the encoded data. In this paper, we consider an $n$-bus large-scale power system with $m$ sensors. The latent distribution is a normal distribution with $n$ variables, corresponding to the state of 118 buses. Notice this does not guarantee that the encoded distribution is the distribution of the system state $\x$. The selection of $n$ is intuitive as we know the sensor data comes from a lower dimensional state space with $n$ variables. In this case, the encoder output is an $n$-dimensional mean vector $\mu_i$ and a $n\times n$ covariance matrix $\Sigma_i$.

A sample $\s_j$ is then drawn from the encoded distribution, where $\s_j\sim\mathcal{N}(\mu_i,\Sigma_i)$. The decoder maps the encoded samples back to the original space of the sensor data to get the reconstructed data $\hat{\z}_i$. The VAE is trained in an unsupervised fashion, where the loss function is a combination of reconstruction loss and regularization loss:
\begin{equation}\label{eq:vaeloss}
    L_{vae}=\E[||\z_i-\hat{\z}_i||^2] + \beta\times\mathbb{KL}(N(\mu_i,\Sigma_i)||N(0,I)),
\end{equation}
where the first term is the sensor data reconstruction loss, and the second term is the regularization loss (scaled by hyperparameter $\beta$ based on the Kullback-Leibler (KL) divergence between the encoded distribution and the prior distribution (which is assumed to be a standard normal distribution). The VAE training algorithm is shown in Algorithm \ref{alg1}.
\begin{algorithm}[!t]
\begin{algorithmic}[1]
\STATE{
\textbf{\textit{input}:} Normal operations sensor datasets ($Z_{0}$), sensor dataset under 3 types of attacks from attack severity 7 $\left(Z_{1,7}^{A}, Z_{2,7}^{A},Z_{3,7}^{A}\right)$, training hyperparameters
}
\STATE{
\textbf{\textit{output}:} Trained VAE under different levels of noise
}
\STATE{
$noise\text{ }levels \leftarrow [0.0,0.05,0.1,0.2,0.5]$
}
\FOR{$level$ in $noise\text{ }levels$}
\STATE{
$\Hat{Z}_{0}\leftarrow$ randomly sample sensor signals from $\left\{Z_{i,7}^{A}\right\}_{i=1}^{3}$ \& add it to $Z_{0}$
}
\STATE{
$\Hat{Z}_{0}^{train},\Hat{Z}_{0}^{test}\leftarrow TrainTestSplit(\Hat{Z}_{0})$
}
\WHILE{ \textit{early-stopping} condition does not satisfied}
\STATE{
Train VAE on $\Hat{Z}_{0}^{train}$ using (\ref{eq:vaeloss}) and evaluate on $\Hat{Z}_{0}^{test}$
}
\ENDWHILE
\ENDFOR
\end{algorithmic}
\caption{Training VAE}
\label{alg1}
\end{algorithm}

\textbf{LSTM:}
RNNs are useful for modeling sequential and time series data. An RNN can be viewed as a special case of the state transition function in (\ref{ngs1}) as the state transitioning distribution $\mathbb{P}(\x_{t}|\x_{t-1},\u_{t-1})$ \cite{rangapuram2018deep,gedon2021deep}. An RNN uses a hidden state as $\mathbf{h}_{t}=f_{\theta}(\mathbf{h}_{t-1}, \u_{t-1})$, and the function parameters $\theta$ are learned using backpropagation through time. Assuming in a ICS the control mechanism is fixed, where $\u_t$ is a function of $\z_t$, we can further modify the RNN to characterize $\mathbf{h}_{t}=g_{\theta}(\mathbf{h}_{t-1}, \z_{t-1})$ Using RNNs provides the one-step ahead expected sensor data. This estimation assists us in attack detection as the attack changes the sensor data that varies from the estimation. After the VAE is trained, the lower dimensional embeddings of the sensor data, (i.e., samples $\{s_{i}\}$ drawn from the encoder) are used to train a one-step look ahead LSTM, which uses $\s_{i-w+1,j},...,\s_{i,j}$ in the time window of size $w$ to predict the future embedding $\s_{i+1}$ at each step $i$. The RNNs characterizes the temporal behavior of the power system based on the lower dimensional embeddings of the sensor data. This allows us to utilize the difference between the prediction and the observed data to identify any abrupt changes in the system behavior caused by cyberattacks. Among the notable types of RNNs, we use LSTM. The input to the LSTM is the embedding at step $t$, $\s_t$. The LSTM mechanism follows
\begin{gather*}
    f_t=sigmoid(W_f\s_t+U_fh_{t-1}+b_f)\\
    i_t=sigmoid(W_i\s_t+U_ih_{t-1}+b_i)\\
    o_t=sigmoid(W_o\s_t+U_oh_{t-1}+b_o)\\
    c_t'=tanh(W_c\s_t+U_ch_{t-1}+b_c)\\
    c_t=f_tc_{t-1}+i_tc_t'\\
    h_t=o_ttanh(c_t)
\end{gather*}
In the above equations, $f_t$ is the forget gate, $i_t$ is the input gate, $o_t$ is the output gate, and $c_t$ is the cell gate, and $h_t$ is the hidden state. The parameters for each gate are represented by $W_\cdot$, $U_\cdot$, and $b_\cdot$. At each step $t$, $\s_{t+1}$ are estimated using a dense layer on the LSTM output $o_t$, where $\hat{\s}_{t+1}=W_d o_t$.
The mean squared error loss is used for training the LSTM:
\begin{equation}\label{eq:rnnloss}
    L_{lstm}=\E[||\s_{i+1}-\hat{\s}_{i+1}||^2],
\end{equation} The training algorithm for LSTM is shown in Algorithm \ref{alg2}

\begin{algorithm}[!t]
\begin{algorithmic}[1]
\STATE{
\textbf{\textit{input}:} Normal operations sensor datasets ($Z_{1}$), sensor dataset under 3 types of attacks from attack severity 7 $\left(Z_{1,7}^{A}, Z_{2,7}^{A},Z_{3,7}^{A}\right)$, time window ($w$), training hyperparameters
}
\STATE{
\textbf{\textit{output}:} Trained RNN under different levels of noise
}
\STATE{
$noise\text{ }levels \leftarrow [0.0,0.05,0.1,0.2,0.5]$
}
\FOR{$level$ in $noise\text{ }levels$}
\STATE{
$\Hat{Z}_{1}\leftarrow$ randomly sample sensor signals from $\left\{Z_{i,7}^{A}\right\}_{i=1}^{3}$ \& add it to $Z_{1}$
}
\STATE{
$S\leftarrow \text{VAE}_{level}.\text{Encoder}(\Hat{Z}_{1})$
}
\STATE{
$S_{\mathbf{X}},S_{\mathbf{y}}\leftarrow$ label time-series lower dimensional embeddings using time window $w$
}
\STATE{
$(S_{\mathbf{X}},S_{\mathbf{y}})^{train},(S_{\mathbf{X}},S_{\mathbf{y}})^{test} \leftarrow TimeSeriesSplit(S_{\mathbf{X}},S_{\mathbf{y}})$
}
\WHILE{ \textit{early-stopping} condition does not satisfied}
\STATE{
Train RNN on $(S_{\mathbf{X}},S_{\mathbf{y}})^{train}$ using (\ref{eq:rnnloss}) and evaluate on $(S_{\mathbf{X}},S_{\mathbf{y}})^{test}$
}
\ENDWHILE
\ENDFOR
\end{algorithmic}
\caption{Training RNN}
\label{alg2}
\end{algorithm}

\textbf{DNN:}
A deep neural network is used for classifying the observations. In order to take both spatio, temporal, and physics information into anomaly detection decision making, we propose a novel data fusion step to generate the input to the DNN, where the VAE reconstruction residual, the LSTM prediction residual $\tilde{\r}_i$, and the state estimation residual are calculated and concatenated to obtain the input to the DNN. Specifically, at step $t$, the current embedding $\s_t$ is decoded into the sensor data space to obtain the reconstruction $\hat{\z}_i$. The VAE residuals is calculated following:
\begin{equation}
    \hat{\r}_i=\hat{\z}_i-\z_i.
\end{equation} This residual will capture spatial correlation among the sensor channels in a data-driven manner.

The predicted embedding $\hat{\s}_t$ from LSTM the previous step is decoded to the original sensor data space by the VAE decoder to obtain the predicted sensor data $\tilde{\z}_i$.
\begin{equation}
    \tilde{\r}_i=\tilde{\z}_i-\z_i.
\end{equation}
This residual will capture the temporal correlation of the sensor data over time.

The state estimation is conducted by solving the least square problem and obtain the measurement estimate $\bar{\z}_i$ from the nonlinear measurement function:
\begin{gather}
    \bar{\x}_i=\text{arg}\min_{\x_i} ||h(\x_i)-\z_i||_2^2\\
    \bar{\z}_i=h(\bar\x_i)
\end{gather} 
The state estimation residuals is calculated following
\begin{equation}
    \bar{\r}_i=\bar{\z}_i-\z_i.
\end{equation} This residual will capture whether the measurement fall into the lower-dimensional state space, which can be viewed as capturing the physics-based correlation.

The three residuals are concatenates to create as the input vector $\r_i=[\hat{\r}_i^T; \tilde{\r}_i^T; \bar{\r}_i^T]$ to the DNN, which is a multilayer perceptron classifier to lable the input data as normal or attack with location information for detection and diagnosis. The DNN is trained on a different dataset than the one that is used to train the VAE and LSTM. Both VAE and LSTM are trained on the normal dataset, while the DNN is trained on a labeled dataset containing both attack data and anomaly data. The cross-entropy loss is used for training the DNN:
\begin{equation}\label{eq:dnnloss}
    L_{dnn} = -\sum_{c=1}^{C} y_{i,c} \log \Hat{y}_{i,c},
\end{equation}
Where $C$ is the number of labels, the true label $y_{i,c}=1$ if $\z_{i}$ is labeled as $c$ and 0 otherwise, and $\Hat{y}_{i,c}$ is the output probability of the DNN. The training algorithm for DNN is shown in algorithm \ref{alg3}

\begin{algorithm}[!t]
\begin{algorithmic}[1]
\STATE{
\textbf{\textit{inputs}:} VAE trained without noise ($\text{VAE}_{0.0}$), RNN trained without noise ($\text{RNN}_{0.0}$), Normal operations sensor data \& state estimation residuals $(Z_{0},R_{0})$, sensor dataset under 3 types of attacks from attack severity 9 $\left((Z_{1},R_{1}), (Z_{2},R_{2}), (Z_{3},R_{3})\right)$, time window ($w$), number of samples ($m$), training hyperparameters
}
\STATE{
\textit{\textbf{output}:} an attack detector
}
\FOR{$i=0,\dots,3$}
\FOR{$t=1,\dots,N$}
\STATE{
$\overline{\mathbf{r}}^{i}_{t} \leftarrow R_{i}[t,:]$
}
\STATE{
$\mathbf{z}_{t}^{i}\leftarrow Z_{i}[t,:]$
}
\STATE{
$count\leftarrow m$
}
\WHILE{$count>0$}
\STATE{
$\Hat{\mathbf{r}}_{t}^{i}\leftarrow \text{VAE}_{0.0}(\mathbf{z}^{i}_{t}) - \mathbf{z}_{t}^{i}$
}
\STATE{
$[\mathbf{s}_{t-w}^{i},...,\mathbf{s}^{i}_{t-1}]\leftarrow \text{VAE}_{0.0}.\text{Encoder}([\mathbf{z}^{i}_{t-w},\dots,\mathbf{z}^{i}_{t-1}])$
}
\STATE{
$\Hat{\mathbf{s}}^{i}_{t}\leftarrow \text{RNN}_{0.0}(\mathbf{s}^{i}_{t-w},...,\mathbf{s}^{i}_{t-1})$
}
\STATE{
$\Tilde{\mathbf{r}}_{t}^{i}\leftarrow \text{VAE}_{0.0}.\text{Decoder}(\Hat{\mathbf{s}}_{t}^{i}) - \mathbf{z}_{t}^{i}$
}
\STATE{
$\mathbf{x}_{t}^{i} \leftarrow [\mathbf{x}_{t}^{i}; [\overline{\mathbf{r}}_{t}^{i}; \Hat{\mathbf{r}}_{t}^{i}; \Tilde{\mathbf{r}}_{t}^{i}]]$
}
\STATE{
$count\leftarrow count-1$
}
\ENDWHILE
\STATE{
$\mathbf{X}^{i}\leftarrow [\mathbf{X}^{i};\mathbf{x}_{t}^{i}]$
}
\ENDFOR
\STATE{
$\mathbf{X} \leftarrow [\mathbf{X};\mathbf{X}^{i}]$
}
\STATE{
$\mathbf{y} \leftarrow [\mathbf{y}; i\times\Vec{\pmb{1}}]$
}
\ENDFOR
\STATE{
$(\mathbf{X},\mathbf{y})^{train},(\mathbf{X},\mathbf{y})^{test} \leftarrow TimeSeriesSplit(\mathbf{X},\mathbf{y})$
}
\WHILE{ \textit{early-stopping} condition does not satisfied}
\STATE{
Train DNN on $(\mathbf{X},\mathbf{y})^{train}$ using (\ref{eq:dnnloss}) and evaluate on $(\mathbf{X},\mathbf{y})^{test}$
}
\ENDWHILE
\end{algorithmic}
\caption{Training attack detector (DNN)}\label{alg3}
\end{algorithm}

\subsection{Online Inference}
\begin{figure}[!ht]
    \centering
    \includegraphics[width=.99\linewidth]{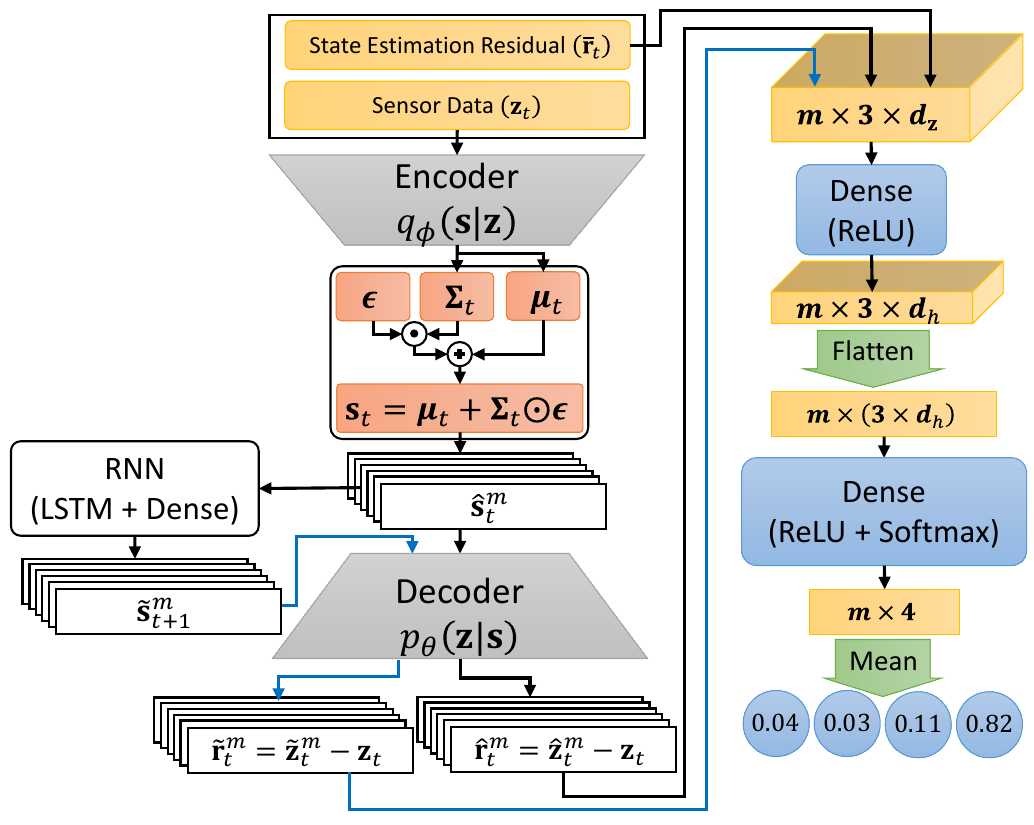}
    \caption{Online inference specifications}
    \label{fig:frm}
\end{figure}

The structure of the online inference framework using the trained VAE, LSTM, and DNN netowrks is shown in Figure \ref{fig:frm}.The input data to the framework is an instance of sensor measurement data at time $t$, denoted as $\z_t$. The VAE encoder first encodes the $\z_t$ vector to the lower dimensional sampling distribution $\N(\mu_t,\Sigma_t)$. Then multiple ($m$) samples are taken from the sampling distribution, denoted as $\s^1_t$,..., $\s^m_t$. These samples are feed into:
\begin{enumerate}
    \item the VAE decoder to compute the VAE residuals $\Hat{\r}_{t}^{1},\dots,\Hat{\r}_{t}^{m}$;
    \item the LSTM to obtain the predicted embeddings; Then, the VAE decoder decodes these predictions to compute the prediction residuals $\Tilde{\r}_{t}^{1},\dots,\Tilde{\r}_{t}^{m}$.
\end{enumerate}
Afterwards, together with the state estimation residuals, $\Bar{\r}_{t}$, these residuals are stacked to form matrices of the form $[\Bar{\r}_{t}^{T};\Hat{\r}_{t}^{{1}^{T}};\Tilde{\r}_{t}^{{1}^{T}}],\dots,[\Bar{\r}_{t}^{T};\Hat{\r}_{t}^{{m}^{T}};\Tilde{\r}_{t}^{{m}^{T}}]$. These matices are then stacked together to form a tensor of the shape $m\times3\times d_{\z}$ as input to the DNN where $d_{\z}$ is the size of the instance of the sensor measurement $\z_{t}$. Figure~\ref{fig:frm} elaborates further on the details of the online inference procedure and the general structure of the proposed DNN. In this structure, at each time $t$, the first block of the dense layers extract the features of each residual signal while reducing their dimensionality. By matricizing the output tensor of the first block, the second block of dense layers extract the features of the combination of residuals to compute the probability of each sample belonging to each class. In other words, the output of the second block is the samples of the classification probability distribution associated with the input sensor measurement $\z_{t}$. Then, the sample mean, as the voting mechanism, determines which class to classify the input sensor measurement $\z_{t}$.

\begin{algorithm}[!t]
\begin{algorithmic}[1]
\STATE{
\textbf{\textit{input}:} Trained attack detector (DNN), VAE trained on one of noise levels, RNN trained on one of noise levels, Sensor data, state estimation residuals, and true labels to infer on $(Z,R,\mathbf{y})$
}
\STATE{
\textbf{\textit{output}:} Attack Detection \& Localization metrics
}
\STATE{
$N\leftarrow length(Z)$
}
\FOR{$t=1,\dots,N$}
\STATE{
$\overline{\mathbf{r}}_{t} \leftarrow R[t,:]$
}
\STATE{
$\mathbf{z}_{t}\leftarrow Z[t,:]$
}
\STATE{
$count\leftarrow m$
}
\WHILE{$count>0$}
\STATE{
$\Hat{\mathbf{r}}_{t}\leftarrow \text{VAE}(\mathbf{z}_{t}) - \mathbf{z}_{t}$
}
\STATE{
$[\mathbf{s}_{t-w},...,\mathbf{s}_{t-1}]\leftarrow \text{VAE}.\text{Encoder}([\mathbf{z}_{t-w},\dots,\mathbf{z}_{t-1}])$
}
\STATE{
$\Hat{\mathbf{s}}_{t}\leftarrow \text{RNN}(\mathbf{s}_{t-w},...,\mathbf{s}_{t-1})$
}
\STATE{
$\Tilde{\mathbf{r}}_{t}\leftarrow \text{VAE}.\text{Decoder}(\Hat{\mathbf{s}}_{t}) - \mathbf{z}_{t}$
}
\STATE{
$\mathbf{x}_{t} \leftarrow [\mathbf{x}_{t}; [\overline{\mathbf{r}}_{t}; \Hat{\mathbf{r}}_{t}; \Tilde{\mathbf{r}}_{t}]]$
}
\STATE{
$count\leftarrow count-1$
}
\ENDWHILE
\STATE{
$\Hat{y}_{t}\leftarrow \text{DNN}(\mathbf{x}_{t})$
}
\STATE{
$\Hat{\mathbf{y}}\leftarrow [\Hat{\mathbf{y}},\Hat{y}_{t}]$
}
\ENDFOR
\STATE{
\textit{DetectionMetric}($\mathbf{y}$, $\Hat{\mathbf{y}}$)
}
\STATE{
\textit{LocalizationMetric}($\mathbf{y}$, $\Hat{\mathbf{y}}$)
}
\end{algorithmic}
\caption{Online Detection \& Localization}
\end{algorithm}

\section{Performance Evaluation}\label{performance}
\subsection{Data Extraction}

In this paper, we consider the IEEE 118-bus model, where the sensor data contains 358 entities, with .... The latent distribution is a normal distribution with 118 variables, corresponding to the state of 118 buses. Notice this does not guarantee that the encoded distribution is the distribution of the system state $\x$. The selection of 118 is intuitive as we know the sensor data comes from a lower dimensional state space with 118 variables. In this case, the encoder output is a 118-dimensional mean vector $\mu_i$ and a $118\times 118$ covariance matrix $\Sigma_i$

We generate data via a simulation study on the IEEE 118-bus model representing a power transmission network and generate the time series of sensor measurements under different conditions. The model contains 118 load buses (substations), 54 generation buses (power generation plants); and 1 slack bus, which is used to balance the active and reactive power in the system and also serves as a reference for all other buses. The model has 186 edges, representing the 186 transmission lines connecting the load and generator buses. The input to the simulation is the load profiles of all the load buses and the power generation plans of all the generation buses. We obtain the load profile of each substation by aggregating the load profiles of a random number of hourly residential power consumption profiles extracted from Pecan Street \cite{street2015dataport}. The generation plan is constructed using the method mentioned in Section \mbox{\ref{sysmodel}}. The simulation in Matlab uses Matpower v7.0~\cite{zimmerman2010matpower} to solve the power flow equation and add measurement noise. The output of the simulation is the hourly time series of the 358 sensor measurements for the simulated period. The 358 sensor measurements include the active power flow on each of the 186 transmission lines, the power generation of each generator bus as well as the slack bus, and the voltage of all the 118 buses.

Recall that the mechanism of a covert attack is to alter the system state by manipulating the control actions. Since in the transmission system we are considering in this simulation, most of the control happens in the power generation plants, we only consider the covert attacks on the generator buses. During the attack, the attacker decreases the generation level by a specific portion. The reason we choose to decrease the generation is from the attacker's objective: compared to generating more power than needed, decreasing the generation will cause possible blackouts and overloading of other generators, which is likely to cause more damage to the system. We simulate the covert attacks on each of the 4 generator buses at 5 levels of severity, where the attacker decreases the power generation by level 1: 10\%, level 2: 20\%, level 3: 30\%, level 4: 40\%, and level 5: 50\% of the planned generation. We assume the attacker obtains access to all the sensors related to the attacked generator and manipulates the sensor measurements by replacing the original values with the ones obtained from simulation such that it shows the attacked generator bus generates the same amount power as planned. For comparison, we simulate the faults as the decrease of power generation by the same amount caused by equipment malfunctioning. The biggest difference between a fault and a covert attack is there is no sensor data manipulation.

\begin{table*}[t!]
\centering
\caption{Overall accuracy, precision, Recall, and F score for DNN classification}
\label{tab:result}
\begin{tblr}{
  cells = {c},
  cell{1}{1} = {c=2}{},
  cell{1}{3} = {c=4}{},
  cell{1}{7} = {c=4}{},
  cell{3}{1} = {r=5}{},
  cell{8}{1} = {r=5}{},
  cell{13}{1} = {r=5}{},
  cell{18}{1} = {r=5}{},
  cell{23}{1} = {r=5}{},
  vline{3} = {-}{},
  vline{2-3,7} = {2-27}{},
  vline{7} = {-}{},
  hline{1-3,8,13,18,23,28} = {-}{},
}
Model  &          & VAE + RNN + DNN   &           &        &         & AE + DNN &           &        &         \\
Noise  & Severity & Accuracy & Precision & Recall & F Score & Accuracy    & Precision & Recall & F Score \\
$0\%$  & 5        & 0.9995   & 0.9995    & 0.9995 & 0.9995  & 0.9996      & 0.9996    & 0.9996 & 0.9996  \\
       & 6        & 0.9997   & 0.9997    & 0.9997 & 0.9997  & 0.9998      & 0.9998    & 0.9998 & 0.9998  \\
       & 7        & 0.9997   & 0.9997    & 0.9997 & 0.9997  & 0.9998      & 0.9998    & 0.9998 & 0.9998  \\
       & 8        & 0.9997   & 0.9997    & 0.9997 & 0.9997  & 0.9997      & 0.9997    & 0.9997 & 0.9997  \\
       & 9        & 0.9993   & 0.9993    & 0.9993 & 0.9993  & 0.9995      & 0.9995    & 0.9995 & 0.9995  \\
$5\%$  & 5        & 0.9998   & 0.9998    & 0.9998 & 0.9998  & 0.9912      & 0.9914    & 0.9912 & 0.9912  \\
       & 6        & 1.0000   & 1.0000    & 1.0000 & 1.0000  & 0.9914      & 0.9916    & 0.9914 & 0.9914  \\
       & 7        & 1.0000   & 1.0000    & 1.0000 & 1.0000  & 0.9914      & 0.9915    & 0.9914 & 0.9913  \\
       & 8        & 1.0000   & 1.0000    & 1.0000 & 1.0000  & 0.9900      & 0.9901    & 0.9900 & 0.9900  \\
       & 9        & 0.9982   & 0.9982    & 0.9982 & 0.9982  & 0.9566      & 0.9579    & 0.9566 & 0.9569  \\
$10\%$ & 5        & 0.9997   & 0.9997    & 0.9997 & 0.9997  & 0.9681      & 0.9714    & 0.9681 & 0.9679  \\
       & 6        & 0.9999   & 0.9999    & 0.9999 & 0.9999  & 0.9658      & 0.9690    & 0.9658 & 0.9657  \\
       & 7        & 0.9999   & 0.9999    & 0.9999 & 0.9999  & 0.9612      & 0.9640    & 0.9612 & 0.9611  \\
       & 8        & 0.9999   & 0.9999    & 0.9999 & 0.9999  & 0.9513      & 0.9527    & 0.9513 & 0.9511  \\
       & 9        & 0.9990   & 0.9990    & 0.9990 & 0.9990  & 0.8150      & 0.8447    & 0.8150 & 0.8162  \\
$20\%$ & 5        & 0.9990   & 0.9990    & 0.9990 & 0.9990  & 0.9745      & 0.9767    & 0.9745 & 0.9744  \\
       & 6        & 0.9992   & 0.9992    & 0.9992 & 0.9992  & 0.9742      & 0.9764    & 0.9742 & 0.9741  \\
       & 7        & 0.9992   & 0.9992    & 0.9992 & 0.9992  & 0.9719      & 0.9738    & 0.9719 & 0.9718  \\
       & 8        & 0.9992   & 0.9992    & 0.9992 & 0.9992  & 0.9614      & 0.9619    & 0.9614 & 0.9613  \\
       & 9        & 0.9987   & 0.9987    & 0.9987 & 0.9987  & 0.7240      & 0.8005    & 0.7240 & 0.7119  \\
$50\%$ & 5        & 0.9993   & 0.9993    & 0.9993 & 0.9993  & 0.9864      & 0.9870    & 0.9864 & 0.9864  \\
       & 6        & 0.9995   & 0.9995    & 0.9995 & 0.9995  & 0.9862      & 0.9867    & 0.9862 & 0.9862  \\
       & 7        & 0.9995   & 0.9995    & 0.9995 & 0.9995  & 0.9783      & 0.9783    & 0.9783 & 0.9782  \\
       & 8        & 0.9995   & 0.9995    & 0.9995 & 0.9995  & 0.9517      & 0.9528    & 0.9517 & 0.9517  \\
       & 9        & 0.9988   & 0.9988    & 0.9988 & 0.9988  & 0.6279      & 0.7539    & 0.6279 & 0.6136  
\end{tblr}
\end{table*}

\begin{figure}[!ht]
    \centering
    \includegraphics[width=.85\linewidth]{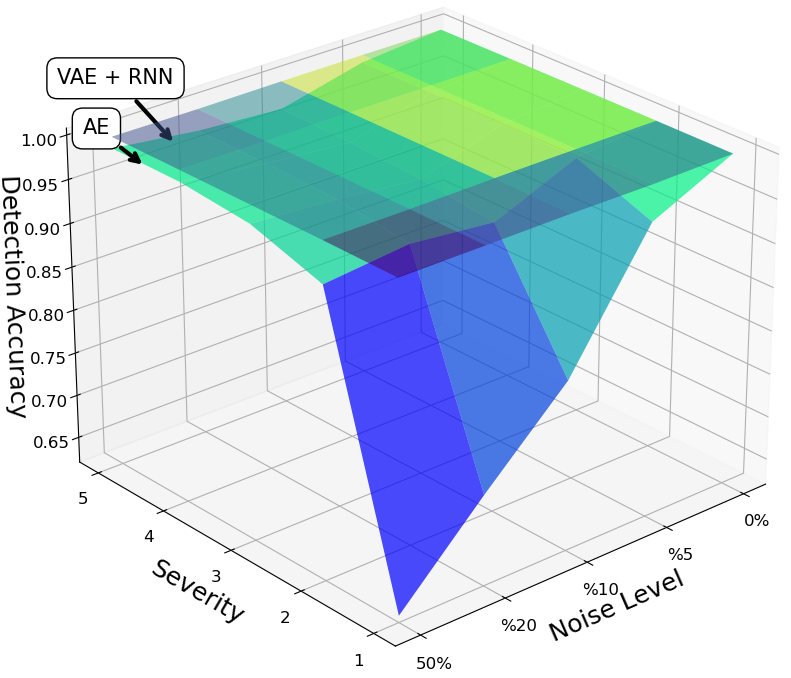}
    \includegraphics[width=.85\linewidth]{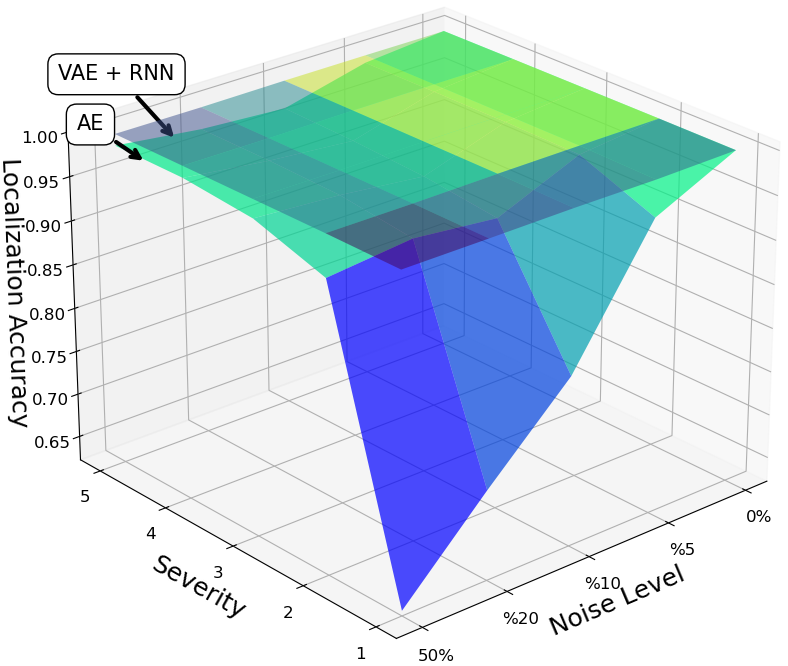}
    \caption{Cyberattack detection and localization accuracy}
    \label{fig:dl}
\end{figure}

\begin{figure*}[ht]
    \centering
    \includegraphics[width=.32\linewidth,valign=t]{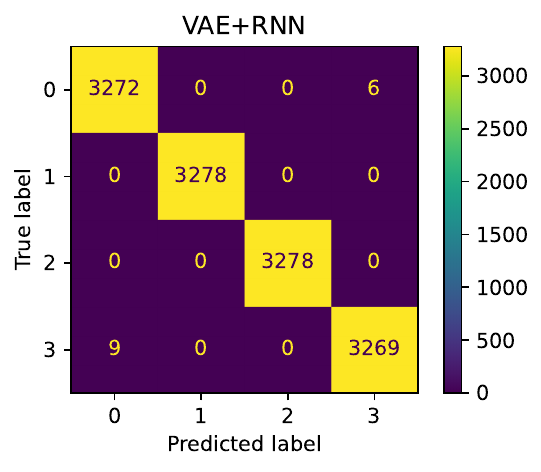}%
    \includegraphics[width=.32\linewidth,valign=t]{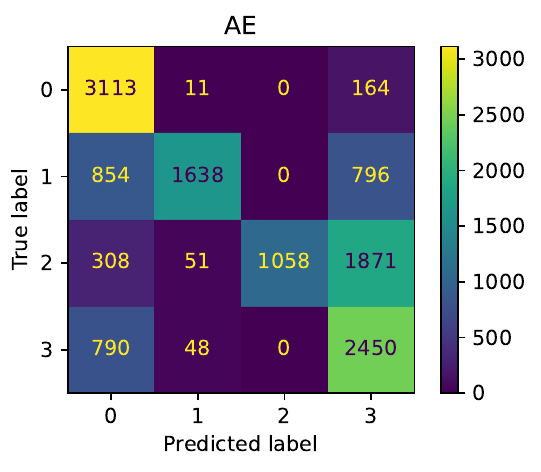}%
    \includegraphics[width=0.25\linewidth,valign=t]{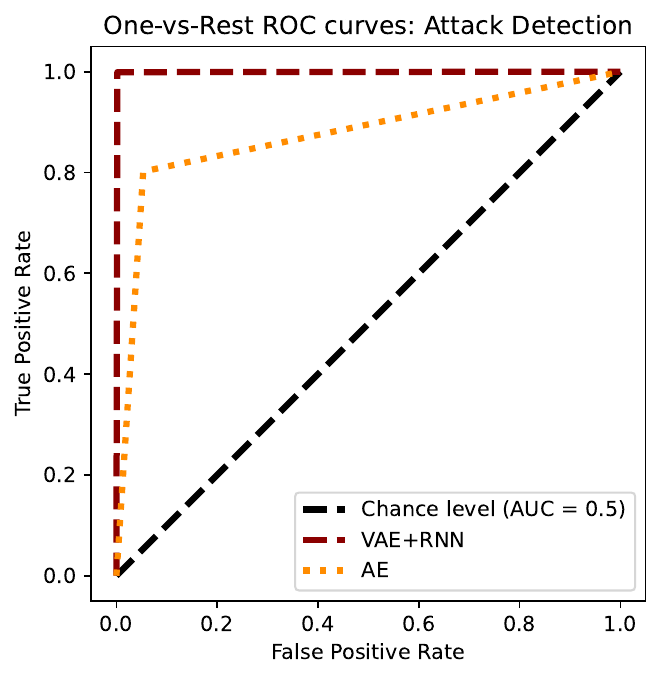}
    \caption{Inference on attack severity 9 dataset with VAE, RNN, and AE trained with $50\%$ noise: confusion matrices for ``VAE + RNN + DNN" (left) and ``AE + DNN" (middle); detection ROC curve (right).}
    \label{fig:950}
\end{figure*}

\subsection{Numerical Study}
\subsubsection{Experiment Design}
We compare our proposed method (VAE + RNN + DNN) with one benchmark to show the advantage of integrating the model-based and data-driven method in recognizing the cyberattacks. This benchmark uses an autoencoder (AE) to project the sensor measurements onto a lower dimensional space that filters out the noise and better represents the system state, which corresponds the steady-state state estimation in traditional frameworks. An AE consists of an encoder to transform the sensor measurements into a latent space, and a decoder to reconstruct the sensor measurements from their latent representation. For generality and having a fair comparison, the code size (output size of the encoder) of AE is chosen as the dimension of the state vector, $n$. In our experiments, both encoder and decoder of AE consists of 2 dense layers with ReLU nonlinearity and normalization layer. We train AE on the same normal dataset as VAE in an unsupervised manner by minimizing the loss function
\begin{equation}
    L_{ae}=\E[||\z_{i}-\check{\z}_{i}||^2],
\end{equation}
After the AE is trained, we calculate the AE residuals using $\check{\r}_{t}=\check{\z}_{t}-\z_{t}$. Then, these residuals are fed into a DNN. This DNN has the same number of dense layers and same nonlinearities as the DNN in our proposed method. However, we drop the flattening and averaging function (see Figure~\ref{fig:frm}).

To show the robustness, we train VAE and RNN, in the proposed model, and AE, in the "AE + DNN" model, on the same normal dataset containing $5\%,10\%,20\%,$ and $50\%$ noise. This noise is randomly selected from attack severity 7 dataset (including randomly choosing the type of attack) to be replaced with the sensor signals in the normal dataset. It is worth mentioning that the DNN in both models are trained using VAE, RNN, and AE trained on normal dataset without any noise. We train the DNN in both models using the attack severity 9 dataset. On inference step, we use the VAE, RNN, and AE trained on noisy datasets to evaluate the detection and diagnosis power of each model.

In both models, in order to train each module, we split the associated dataset into train-test sets with $80\%-20\%$ proportion while preserving the temporal nature of the time series. To avoid overfitting, we use early-stopping regularization.

\subsubsection{Results}
The performance of the method is evaluated by accuracy, precision, recall, and the F1-score, which are shown in Tab.\ref{tab:result} that shows the overall classification performance of DNN among the 4 labels (normal and 3 attack types) in each attack severity dataset. The predicted labels are computed as 
\begin{equation*}
    \Hat{c}\leftarrow\arg\max_{c\in C} \Hat{y}_{t,c},
\end{equation*}

Table~\ref{tab:result} and Figure~\ref{fig:dl} denote the overall localization and detection accuracy of the proposed model and the ``AE + DNN" model evaluating on different levels of attacks. We observe that the value of all metrics increases as the level (severity) of the attack increases. This is a validation that the covertness (the ability to stay undetected) of the attack decreases as the severity of attack increases, meaning the distinction between normal data and the data under attack becomes clearer as the severity of attack increases, leading to a higher detection power and diagnosis accuracy. However, as the percentage of noise increases while training these modules, all metrics almost remain consistent as opposed to the ``AE + DNN" model. As we include more noise into the dataset while training AE offline, the reconstructed sensor measurements (and, accordingly, the residuals) tend to mislead the downstream DNN. This observation is because AE extracts only the spatial correlation of sensor measurements that is deterministic for each time. This is a validation that the latent distribution $\mathcal{N}(\mu_t,\Sigma_t)$ learned by VAE is robust to the noises and, together with RNN, the proposed model captures both spatial and temporal correlations of the sensor measurements effectively. Furthermore, the estate estimation residuals bring more robustness and effectiveness by incorporating system physics. 

From Table~\ref{tab:result} and Figure~\ref{fig:dl}, we observe that ``AE + DNN" has significantly low detection and localization accuracy when it is being tested on the attack severity 9 dataset and AE is trained with $50\%$ noise. Figure~\ref{fig:950} shows the confusion matrices of the proposed model when both VAE and RNN are trained with $50\%$ noise vs. the ``AE + DNN" model AE is trained with $50\%$ noise while doing inference on attack severity 9 dataset. We observe the ``AE + DNN" model cannot recognize each attack type; Moreover, it recognizes attacks as normal sensor signals while classifying some normal sensor signals as attack. The ROC curve in Figure~\ref{fig:950} elaborates more on the diagnostic ability of both models. We observe that the proposed model has higher area under curve (AUC) score than the ``AE + DNN" model that indicates the proposed model has a good measure of separability.

\section{Conclusion}\label{concl}
In this paper, we proposed a generic data-driven framework for detecting, diagnosing, and localizing covert attacks on industrial control systems. The proposed framework uses an autoencoder for unsupervised feature extraction from sensor measurements and then uses an RNN to capture the temporal correlations among the encoded sensor measurements. The prediction of the RNN is decoded and compared with the input sensor measurements to get the residuals, which help detect the anomaly. The residuals and the sensor measurements are processed with a DNN to determine whether an observation represents normal conditions, an attack, or a fault, and to identify the location of the attack/fault. The proposed framework was compared with the model-based state estimation technique, as well as a modification of itself by removing the autoencoder. The results showed that the autoencoder helps extract the important features from the data, as well as reduce the dimension of the input to the RNN. This significantly helps improve the classification accuracy of the DNN. It is worth noticing that the RNN does not provide a more accurate estimation of the state compared to the model-based state estimation. However, since the RNN considers the temporal behavior of the system, which is not considered by the mode-based SE, the residuals obtained from the decoded RNN prediction could better capture the anomalous characteristics of the data when the system is under attacks/faults. The reason model-based SE does not perform well under covert attack is that the objective of SE is to minimize the residuals. This leverages the estimation of the state to normal conditions, which does not represent the underlying truth, especially when the attacker has access to more sensors. The simulation study and performance evaluation validated the proposed method. Since this method is model-free, it is easily generalizable to other networked industrial control systems.

In this work, we only trained and tested the model on the known attack/fault types. A future direction is to extend the method to anomaly-based detection, which can detect novel attacks and faults. Another direction is to combine the method with graphical network topology and the correlation structure of the data.

\normalsize
\bibliography{references}


\end{document}